\documentclass[10pt,twocolumn,letterpaper]{article}

\usepackage[pagenumbers]{cvpr}

\usepackage[dvipsnames]{xcolor}

\definecolor{cvprblue}{rgb}{0.21,0.49,0.74}
\usepackage[pagebackref,breaklinks,colorlinks,allcolors=cvprblue]{hyperref}
\usepackage{multirow}
\usepackage{amsmath}
\usepackage[numbers,sort&compress]{natbib}
\usepackage{makecell}
\usepackage{graphicx}
\usepackage{sidecap}

\begin{document}

\title{NeuroPump: Simultaneous Geometric and Color Rectification for Underwater Images}

\author{Yue Guo$^{1}$\qquad  Haoxiang Liao$^{1}$\qquad  Haibin Ling$^{2}$\qquad  Bingyao Huang$^{1}$\footnote{}\vspace{6pt}\\
$^{1}$ Southwest University\qquad 
$^{2}$ Stony Brook University\\
{\tt\small \{guoyue, hliao\}@email.swu.edu.cn}\qquad 
{\tt\small hling@cs.stonybrook.edu}\qquad 
{\tt\small bhuang@swu.edu.cn}
}

\twocolumn[{%
\renewcommand\twocolumn[1][]{#1}%
\maketitle

\begin{center}
    \centering
    \captionsetup{type=figure}
    
    \includegraphics[width=1.\textwidth]{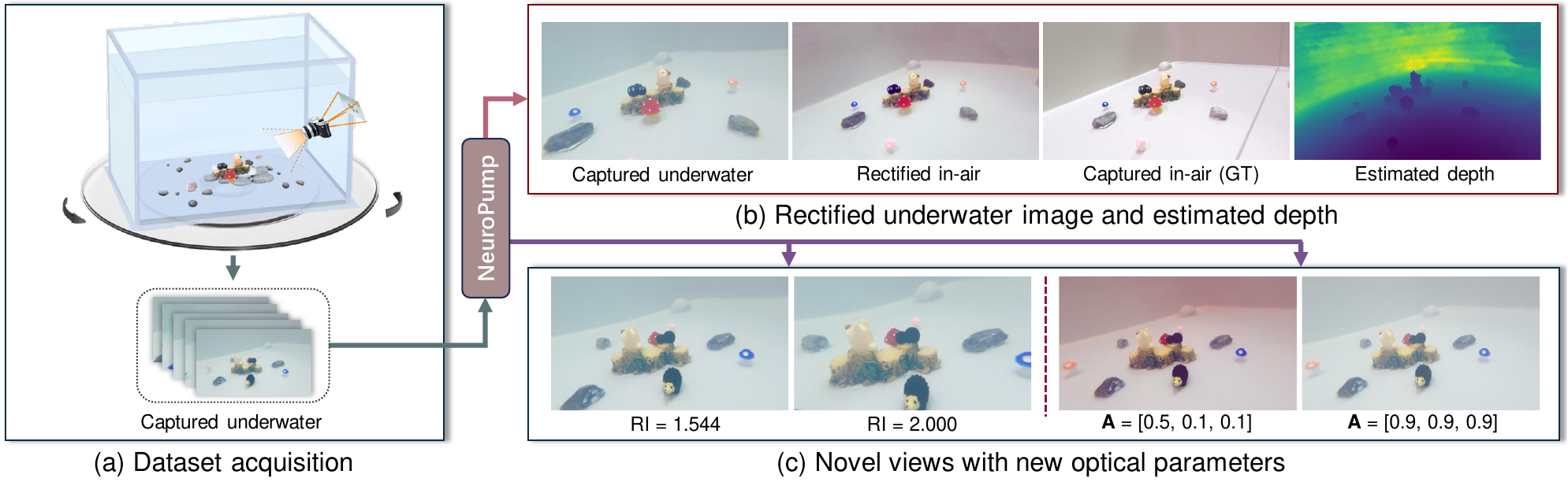} 
    \vspace{-4.5mm}
    \caption{NeuroPump aims to simultaneously rectify underwater images' geometric and color distortions and estimate depth (b) from (a) multiple underwater views, just like the situation (the 3rd in (b)) where water is pumped out. (c) After training, NeuroPump can synthesize novel views by controlling the camera pose, medium refractive indices (RI) and global background light $\textbf{A}$.}
    \label{fig:teaser}
\end{center}%
}]
\maketitle

\begin{abstract}
\vspace{-3mm}
Underwater image restoration aims to remove geometric and color distortions due to water refraction, absorption and scattering.
Previous studies focus on restoring either color or the geometry, but to our best knowledge, not both. However, in practice it may be cumbersome to address the two rectifications one-by-one.
In this paper, we propose NeuroPump, a self-supervised method to simultaneously optimize and rectify underwater geometry and color as if water were pumped out.
The key idea is to explicitly model refraction, absorption and scattering in Neural Radiance Field (NeRF) pipeline, such that it not only performs simultaneous geometric and color rectification, but also enables to synthesize novel views and optical effects by controlling the decoupled parameters. 
In addition, to address issue of lack of real paired ground truth images, we propose an underwater 360 benchmark dataset that has real paired (\ie, with and without water) images. Our method clearly outperforms other baselines both quantitatively and qualitatively. 
Our project page is available at: \href{https://ygswu.github.io/NeuroPump.github.io/}{https://ygswu.github.io/NeuroPump.github.io/}.
\end{abstract}

\section{Introduction}
\label{sec:intro}

Underwater images suffer from geometric and color distortions caused by the complex underwater environment, leading to low contrast, color cast, and distorted geometry. These distortions may hinder underwater applications. 

To address these issues, previous works either focus on rectifying the color distortion \cite{peng2018generalization, he2010single, chiang2011underwater, song2018rapid, li2016single, akkaynak2019sea, li2017watergan, li2019underwater, li2020underwater, li2018emerging} or geometric distortion (\eg, refraction) \cite{kang2012experimental, jordt2012refractive, jordt2013refractive, lavest2000underwater}. However, many underwater applications require both geometric and color rectification, such as underwater oceanography \cite{li2022advanced}, archaeology \cite{drap2015rov}, navigation enhancement \cite{lee2012vision}, underwater remote sensing \cite{liu2020uwdetecting} and underwater virtual reality \cite{costa2017towards, blum2009augmented, morales2009underwater}. Although a dome port \cite{levy2023seathru,venkatramanan2022waternerf} may be used to mitigate geometric distortion, they may suffer from buoyancy, large size and glare, and are less flexible in practice.

Another challenge of underwater imaging is the lack of real captured in-air ground truth for training and evaluation.
Intuitively, it seems easy to construct a lab environment to capture paired images before and after adding water. However, due to subtle water tank deformation after filling water and suspended objects, bubbles and underwater illumination changes, the camera-captured image pairs may be geometrically and photometrically misaligned. This issue deteriorates when performing 360 multi-view shooting. 
Therefore, most previous benchmarks use synthetic paired datasets (\ie, with and without water) \cite{li2019underwater, li2020underwater} for model training and evaluation. However, it is nontrivial for synthetic data to consider all physical properties, such as \textbf{geometric distortion} due to refraction, \textbf{real global illumination} and \textbf{scattering}. On the other hand, without real paired ground truth images, it is hard to accurately evaluate underwater image restoration methods. Although some ground truth-free metrics were proposed, such as UCIQE \cite{yang2015underwater}, UIQM \cite{panetta2015human}, image entropy and visible edges \cite{hautiere2008blind}, they may not match exactly human perception \cite{li2019underwater}.

In this paper, we propose NeuroPump, a self-supervised method to simultaneously rectify underwater geometric and color distortions, and to synthesize novel views and novel underwater properties (\autoref{fig:teaser}) for the trained scene. The key idea is to explicitly model refraction (geometric distortion), scattering and absorption (color distortion) in the NeRF pipeline \cite{mildenhall2021nerf, barron2022mip}. 

In particular, for geometric distortion, NeuroPump explicitly models ray refraction at the lens case interface between the camera lens and the water, and bends the rays according to the Snell's law~\cite{hecht2002optics} to rectify the scene geometry (depth). For color distortion (\ie, scattering and absorption), as underwater distorted color mainly consists of the attenuated direct radiance and back-scatter~\cite{jaffe1990computer}, we explicitly learn global background light, attenuation coefficients and unattenuated direct radiance, from which we also calculate the radiance intensity compensation factor to rectify the color of underwater scene. As parameters above are decoupled, it allows not only simultaneous geometric and color rectification, but also novel view synthesis with new optical parameters.

Moreover, we propose a real captured underwater 360-degree view dataset that has paired (\ie, with and without water) ground truth for accurate model evaluation. Our dataset consists of five setups and each setup has around 60 views. Its underwater images have both color and geometric distortions, which always emerge when we capture underwater scenes using a planar lens camera.
As mentioned before, this task is challenging, since it is almost impossible to pump out the water in the natural environment to capture accurately aligned in-air images. Therefore, following the convention in general image enhancement~\cite{ancuti2018haze, filin2023single, Chen2024dehazenerf,skinner2017automatic, kohler2012recording,zingg2010mav, fei2013comprehensive, wang2023neref, thapa2020dynamic}, 
we start with lab environment first, where the conditions are easier to control, and are still more realistic than synthetic data.

Our contributions can be summarized as follows: 
\begin{itemize}
    \item Our NeuroPump is the first to simultaneously rectify geometric and color distortions of underwater images.
    \item By explicitly decoupling and learning the physical factors, NeuroPump can synthesize novel views and novel refraction, absorption and scattering effects.
    \item We construct the first underwater 360 benchmark dataset that has real paired (\ie, with and without water) ground truth images for accurate model evaluation, and our method clearly outperforms previous arts.
\end{itemize}

\begin{figure*}[!t]
  \centering
   \includegraphics[width=1.0\linewidth]{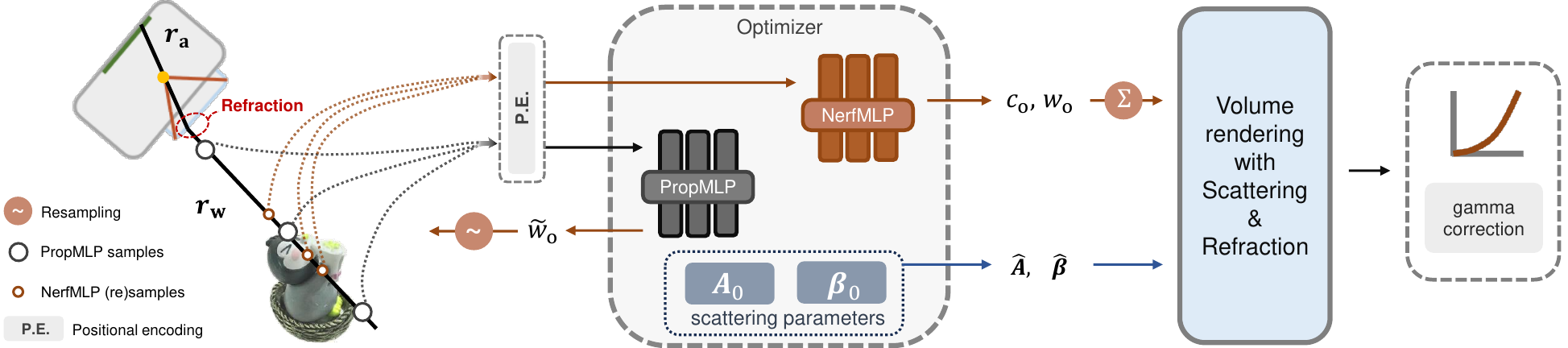}
   \caption{NeuroPump pipeline. NeuroPump begins by applying the Snell's law to mitigate refraction in ray sampling. First, PropMLP takes coarse samples to get approximate object weight $\tilde{w}_{\textit{o}}$. Next, $\tilde{w}_{\textit{o}}$ is used to take NerfMLP resamples to obtain fine object color $c_{\textit{o}}$ and weight $w_{\textit{o}}$.
   Two scattering parameters, namely the global background light $\mathbf{A}$ and the water attenuation coefficients $\boldsymbol{\beta}$, are trained together with MLPs. $\mathbf{A_0}$ and $\boldsymbol{\beta_0}$ represent the initial values. Third, gamma correction is applied after volume rendering, to convert from linear RGB space to sRGB space. All parameters are optimized mainly by minimizing the loss between the model inferred and camera-captured multi-view underwater images.}
   \label{fig:flowchart}
\end{figure*}

\vspace{-1mm}
\section{Related work}

\subsection{Underwater imaging}
\vspace{-1mm}
\noindent\textbf{Underwater geometry rectification} mainly focuses on rectifying geometric distortion due to refraction, and has been studied in ocean remote sensing \cite{liu2020uwdetecting}, underwater reconstruction \cite{kang2012experimental} and underwater surveying \cite{brandou20073d}.
Lavest \etal \cite{lavest2000underwater} discussed the relationship between radial distortion and water refraction distortion, and proposed to approximate the refraction distortion using radial distortion. But it was not physics-based and 
only worked for cameras without lens case \cite{jordt2012refractive, jordt2013refractive}. Jordt-Sedlazeck \etal \cite{jordt2012refractive} presented an underwater camera calibration technique by considering the distance between the camera optical center and the camera lens case. This method integrated an evolutionary optimization algorithm with synthetic analysis. Based on this work, Jordt-Sedlazeck \etal further introduced a refraction plane scanning approach for dense underwater 3D reconstruction \cite{jordt2013refractive}. However, this approach relies on good initialization. Chadebecq \etal \cite{chadebecq2017refractive} studied an approach to infer camera intrinsics and extrinsics using underwater images with significant refraction by assuming a thin camera lens case.

\noindent\textbf{Underwater color rectification} mainly addresses color cast, blur and low contrast. Previous works can be divided into physics-based and supervised learning-based methods.
Classic physics-based methods explicitly formulate scattering and absorption using depth and ambient light, and are widely used in dehazing \cite{nayar1999vision} and underwater image restoration \cite{jaffe1990computer}. The difference is that underwater image restoration assumes that each color channel has a different attenuation factor. Physics-based methods \cite{peng2018generalization, chiang2011underwater} introduced dark channel prior \cite{he2010single} into underwater imaging to obtain the scene depth, global background light and attenuation coefficients. Li \etal \cite{li2016single} rectified green and blue channels by dark channel prior and red channel by the Gray-World assumption. However, these models may not generalize well across diverse waters. Akkaynak and Treibitz~\cite{akkaynak2018revised} proposed a revised formulation to address the degradation disparity between the unattenuated direct radiance and back-scatter, and improved the generalization of previous work. Despite the effort, simultaneously estimating unknown depth, attenuation coefficients and global background light remains challenging for pure physical-based methods, due to limited model parameters and handcrafted priors.

Supervised learning-based methods such as WaterGAN \cite{li2017watergan}, Water-Net \cite{li2019underwater} and UWCNNs \cite{li2020underwater} can learn complex scene priors and optics from the training data, and generally outperform classic physics-based ones. Li \etal \cite{li2018emerging} produced a weakly supervised color transfer model (\ie, Water CycleGAN) to rectify underwater image color distortion. Jiang \etal proposed a light weight model F$\text{A}^{+}$Net \cite{jiang2023five} with only 9k parameters, enabling real-time underwater image enhancement. However, supervised methods require a large amount of paired training data (\ie, with and without water). Moreover, single-view depth estimation is usually ambiguous, and the results are less accurate compared with multi-view approaches, \eg, a self-supervised method by Nisha \etal \cite{varghese2023self} and NeRF-based \cite{levy2023seathru}.

\subsection{NeRF-based methods}

NeRF-based approaches~\cite{huang2022hdr,jun2022hdr,mildenhall2022nerf,cui2023aleth,guo2022nerfren,verbin2022ref,bemana2022eikonal} showed promising results in dehazing \cite{ramazzina2023scatternerf, jin2023reliable, Chen2024dehazenerf}, underwater image restoration and novel view synthesis. 
ScatterNeRF \cite{ramazzina2023scatternerf} and DehazeNeRF \cite{Chen2024dehazenerf} extended NeRF with atmospheric scattering for reconstructing scenes with low visibility due to haze, and rectifying color and geometry for hazed scenes.
WaterNeRF \cite{venkatramanan2022waternerf} applied NeRF to histogram equalized underwater images for color rectification. 
SeaThru-NeRF \cite{levy2023seathru} designed a multilayer perceptron to infer back-scatter and attenuation coefficients to address underwater color distortion. 
Neural Underwater Scene Representation \cite{tang2024uwnerf} applied 2 extended MLP for dynamic objects and unstable illumination field.
Despite the effort on color rectification and novel view synthesis, existing approaches do not address geometric and color distortions simultaneously. Therefore, we proposed NeuroPump to achieve joint optimization and simultaneous rectification of color and geometric distortions.

\subsection{Underwater datasets}
Several real captured underwater datasets have been proposed for model evaluation \cite{duarte2016dataset, akkaynak2019sea, xiao2010sun, berman2020underwater}. Some of them provide only underwater single-view images \textit{without} in-air ground truth, such as underwater images in the SUN dataset \cite{xiao2010sun} and those by Duarte \etal \cite{duarte2016dataset} with different degradation by milk, chlorophyll and green tea in a water tank. Other underwater datasets provide measured depth information, \eg, Sea-thru \cite{akkaynak2019sea} consists of 1,100 underwater images with range maps. Haze-line \cite{berman2020underwater} captures raw underwater images and distance maps. However, most real captured underwater datasets above provide only underwater images and do not include their corresponding in-air ground truth. Because pumping out the water is extremely challenging, not to mention serious lighting and alignment requirements.

Due to the equipment and environment constraints, obtaining an underwater dataset with paired ground truth is challenging. Therefore, synthetic datasets were studied. Li \etal \cite{li2019underwater} collected a diverse set of real underwater images, and applied different color rectification methods to obtain their in-air appearance. The best results were chosen as the approximating in-air ground truth.
Li \etal \cite{li2020underwater} rendered underwater effects of a set of in-door RGB-D images, including ten subsets for different types of water. Ye \etal \cite{ye2022underwater} generated a large underwater dataset (LNRUD) using a neural rendering model, which learned the natural degradation process. However, these synthetic data may not fully account for all real-world optical properties, such as refraction and global illumination consistency.

\begin{figure*}[t]
\centering
  \includegraphics[width=.98\linewidth]{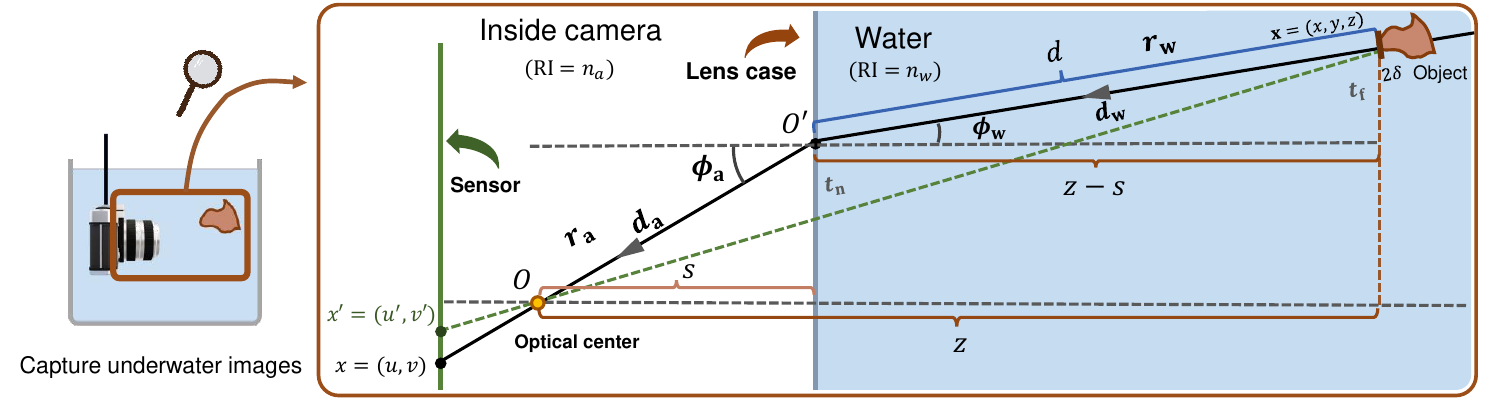}
\caption{Underwater imaging process. Right box is the zoom-in of Left box. Let $o$ be the camera optical center, $s$ and $z-s$ denote the perpendicular distances between the camera optical center and object point to the lens case interface, respectively. An underwater light ray $\mathbf{r}_{\text{w}}$ with direction $\mathbf{d}_\text{w}$ first refracts at the camera lens case interface, and travels along direction  $\mathbf{d}_\text{a}$, and finally passes through the camera optical center and hits the sensor at pixel $\boldsymbol{x}$. The angle of incidence and the angle of refraction are $\phi_{\text{w}}$ and $\phi_{\text{a}}$, respectively. Note that the directions of $\mathbf{d}_\text{w}$ and $\mathbf{d}_{\text{a}}$ are reversed in NeuroPump pipeline in \autoref{sec:method}.}
\label{fig:geo_ref}
\end{figure*}

\section{NeuroPump}
\label{sec:method}

\subsection{Refraction formulation with Snell's Law }\label{sec:refraction} 
\vspace{-0.5mm}
Underwater imaging process involves refraction, absorption, and scattering, leading to both geometric and color distortions. 
An important module of NeuroPump is the modeling of underwater refraction, namely, how to establish the connection between underwater \textit{refracted} ray $\mathbf{r}_{\text{w}}$ and non-refracted ray $\mathbf{r}_{\text{a}}$ in air. Below, we show how we incorporate the Snell's law \cite{hecht2002optics} to our formulation.

As shown in \autoref{fig:geo_ref}, given the camera optical center $\mathbf{o}$ and a pixel coordinated $\boldsymbol{x}=(u,v)$, the origin of the refracted ray $\mathbf{o'}(\boldsymbol{x})$ is determined by:
\begin{equation}
\small
\begin{aligned}
  \mathbf{o'}(\boldsymbol{x}) = \mathbf{o} + \frac{s}{{\cos(\phi_{\text{a}}(\boldsymbol{x}))}} \mathbf{d}_{\text{a}}(\boldsymbol{x})~~,
\end{aligned}
\label{eq:16+}
\end{equation}
where $s$ is the perpendicular distance between the camera optical center and lens case interface; $\mathbf{d_{\text{a}}}(\boldsymbol{x})$ is the direction of the ray that passes through $\mathbf{o}$ and pixel $\boldsymbol{x}$; and  $\phi_{\text{a}}$ is the angle of refraction, which is given by the Snell's law:
\begin{equation}
\small
  \phi_{\text{a}}(\boldsymbol{x}) = \arcsin\left( \frac{n_{\text{w}}}{n_{\text{a}}}\sin\left( \phi_{\text{w}}(\boldsymbol{x}) \right) \right)~,
\label{eq:4}
\end{equation}
where $n_{\text{a}}$ and $n_{\text{w}}$ denote the refractive indices of the medium inside and outside the lens case, respectively. We use Rodrigues' rotation formula to rotate $\mathbf{d}_{\text{a}}$ to $\mathbf{d}_{\text{w}}$ (see Supp. for details), and the refracted ray is given by:
\begin{equation}
\small
\begin{aligned}
  \mathbf{r}_{\text{w}}\left( \boldsymbol{x} \right) = \mathbf{o'}(\boldsymbol{x}) + t~ \mathbf{d}_{\text{w}}(\boldsymbol{x})~;~~
  \mathbf{d}_{\text{w}} = \textbf{Refract}\left( {\mathbf{d}_{\text{a}},\phi_{\text{a}},\phi_{\text{w}}} \right)~,
\end{aligned}
\label{eq:16++}
\end{equation}
where $t$ represents the distance along $\mathbf{r}_{\text{w}}$.

\subsection{Absorption/scattering formulation}
\label{sec:problem_formulation}
It is worth noting that water scattering consists of forward-scattering and back-scattering. Forward-scattering is only obvious in extremely turbid waters, where most NeRF-based methods that use COLMAP may fail due to unsuccessful keypoint matching for camera pose estimation. Therefore, NeuroPump focus on back-scattering that occurs in less turbid waters.

As shown in \autoref{fig:geo_ref}, the \textit{refracted} ray is $\mathbf{r}_{\text{w}}=\mathbf{o'}+t\mathbf{d}_{\text{w}}$, whose origin is $\mathbf{o'}$ and direction is $\mathbf{d}_{\text{w}}$; denote the global background light as $\mathbf{A}$ and the medium attenuation coefficients as $\boldsymbol{\beta}$ (for absorption/scattering due to water and suspended particles), the camera-captured radiance $I\left( \mathbf{r}_{\text{w}} \right)$ is expressed as:
\begin{equation}
\small
\begin{aligned}
    &I\left( \mathbf{r}_{\text{w}} \right) = I_{\text{o}}\left( \mathbf{r}_{\text{w}} \right) + I_{\text{m}}\left( \mathbf{r}_{\text{w}} \right) \\& =
   {\int_{t_\text{n}}^{t_\text{f}} \!\! {T(t) {\sigma_{\text{o}}(t)\mathbf{c}_{\text{o}}\left( {t,\mathbf{d}_{\text{w}}} \right)} dt}} + \!\!{\int_{t_\text{n}}^{t_\text{f}} \!\! {T(t){\boldsymbol{\beta}(t)\mathbf{A}\left( t,\mathbf{d}_{\text{w}} \right)}dt}},
\end{aligned}
\label{eq:PF1}
\end{equation}
where $I_{\text{o}}\left( \mathbf{r}_{\text{w}} \right)$ is underwater object radiance and  $I_{\text{m}}\left( \mathbf{r}_{\text{w}} \right)$ is the radiance brought by water and suspended particles.  $t_\text{n}$  and  $t_\text{f}$ are respectively the near and far ray bounds.
$\sigma_\text{o}$ and $\mathbf{c}_\text{o}$ are respectively object density and color. $T(t)$ is the transmittance of the sampled field at position $\mathbf{r}_{\text{w}}(t)$ given by:
\begin{equation}
\small
\begin{aligned}
   T(t) \!=\! T_\text{o}(t)T_\text{m}(t) = {\exp\Big(\! {- \!{\int_{t_\text{n}}^{t}\!\!{\sigma_\text{o}(l)dl}}} \Big)}{\exp\Big(\! {-\! {\int_{t_\text{n}}^{t}\!\! {\boldsymbol{\beta}(l)dl}}} \Big)}.
\end{aligned}
\label{eq:PF2}
\end{equation}
$T(t)$ can be decomposed into two components: the absorption $T_\text{o}(t)$ due to the object and the absorption/scattering $T_\text{m}(t)$ due to the water and suspended particles.
We define the distance from ray origin to the nearest object point on the ray as $d$. The radiance of underwater objects $I_{\text{o}}\left( \mathbf{r}_{\text{w}} \right)$ is accumulated along the ray, and we divide accumulation into three parts: $\lbrack t_\text{n},d-\delta)$; $\lbrack d-\delta,d+\delta \rbrack$; and $(d+\delta, t_\text{f}\rbrack$; where $\delta \to 0$ is a very small distance. Thus, the accumulation at $\mathbf{r}_{\text{w}}(d)$ can be approximated by:
\begin{flalign}
\small
\begin{aligned}   
   I_{\text{o}}\left( \mathbf{r}_{\text{w}} \right) \approx &{\int_{t_{\text{n}}~}^{d-\delta}{T_{\text{m}}(t)T_{\text{o}}(t) {\sigma_{\text{o}}(t)\mathbf{c}_{\text{o}}\left( {t,\mathbf{d}_{\text{w}}} \right)} dt}} \\
   &+ T_{\text{m}}(d)T_{\text{o}}(d)\sigma_{\text{o}}(d)\mathbf{c}_{\text{o}}\left( {d,\mathbf{d}_{\text{w}}} \right)  \\
   &+ {\int_{d+\delta~}^{t_{\text{f}}}T_{\text{m}}(t){T_{\text{o}}(t){\sigma_{\text{o}}(t)\mathbf{c}_{\text{o}}\left( {t,\mathbf{d}_{\text{w}}} \right)} dt}}~.\\
\end{aligned}
\label{eq:PF3}
\end{flalign}
Assuming the underwater objects are opaque, \ie, $\sigma_\text{o}(d) \approx 1$ and the transmittance $T(t)\approx T_{\text{o}}(t)\approx 0$ for $t\in( d+\delta, t_\text{f}\rbrack$. Additionally, according to the definition of $d$, there are no objects in $\lbrack t_\text{n},d-\delta)$, thus $\sigma_{\text{o}}\approx 0$. Then, considering $d$ is unknown before well-sampling and training, we have:
\begin{flalign}
\small
\begin{aligned}  
   I_{\text{o}}\left( \mathbf{r}_{\text{w}} \right) & \approx T_{\text{m}}(d)T_{\text{o}}(d)\sigma_{\text{o}}(d)\mathbf{c}_{\text{o}}\left( {d,\mathbf{d}_{\text{w}}} \right),
\end{aligned}
\label{eq:PF3+}
\end{flalign}
where the $d$ in $T_{\text{m}}(d)$ is decided by the first sample point of the opaque object in the training ray after each iteration.

Similarly, $I_{\text{m}}\left( \mathbf{r}_{\text{w}} \right)$ can be divided into three intervals:
\begin{flalign}
\small
\begin{aligned}
   I_{\text{m}}\left( \mathbf{r}_{\text{w}} \right) = &{\int_{t_{\text{n}}~}^{d-\delta}{T_{\text{m}}(t)T_{\text{o}}(t){\boldsymbol{\beta}(t)\mathbf{A}\left( t,\mathbf{d}_{\text{w}} \right)}dt}} \\
   &+ T_{\text{m}}(d)T_{\text{o}}(d)\boldsymbol{\beta}(t)\mathbf{A}\left( {t,\mathbf{d}_{\text{w}}} \right) \\
   &+ {\int_{d+\delta~}^{t_{\text{f}}}{T_{\text{m}}(t)T_{\text{o}}(t){\boldsymbol{\beta}(t)\mathbf{A}\left( t,\mathbf{d}_{\text{w}} \right)}dt}}~. \\
\end{aligned}
\label{eq:PF4}
\end{flalign}
Since water density $\boldsymbol{\beta}(t)\approx 0$ in $(d+\sigma, t_{\text{f}}\rbrack$ and $\lbrack d-\sigma,d+\sigma \rbrack$, and the radiance behind the opaque object point (\ie, in the interval $( d+\delta, t_\text{f}\rbrack$) is occluded, consequently, the radiance contributed by water and suspended particles in $(d+\sigma, t_{\text{f}}\rbrack$ and $\lbrack d-\sigma,d+\sigma \rbrack$ is approximately 0, and we have:
\begin{flalign}
\small
   I_{\text{m}}\left( \mathbf{r}_{\text{w}} \right) \approx  {\int_{t_\text{n}~}^{d}{T_{\text{m}}(t) {\boldsymbol{\beta}(t)\mathbf{A}\left( {t,\mathbf{d}_{\text{w}}} \right)} dt}}~.
\label{eq:PF4}
\end{flalign}
Thus, $I\left( \mathbf{r}_{\text{w}} \right)$ becomes:
\begin{equation}
\small
\begin{aligned}
   I\left( \mathbf{r}_{\text{w}} \right) \approx 
 &T_{\text{m}}(d){\int_{t_\text{n}~}^{t_\text{f}}{T_{\text{o}}(t){\sigma_{\text{o}}(t)\mathbf{c}_{\text{o}}\left( {t,\mathbf{d}_{\text{w}}} \right)} dt}} + \\&
 {\int_{t_\text{n}~}^{d}{T_{\text{m}}(t){\boldsymbol{\beta}(t)\mathbf{A}\left( {t,\mathbf{d}_{\text{w}}} \right)} dt}}~.\\
\end{aligned}
\label{eq:I_w}
\end{equation}
Denote the unattenuated underwater direct radiance as $J\left( \mathbf{r}_{\text{w}} \right) = {\int_{t_\text{n}}^{t_\text{f}}{{T}_\text{o}(t){\sigma_\text{o}(t)\mathbf{c}_\text{o}\left( {t,\mathbf{d}_{\text{w}}} \right)} }}dt$, 
and assume the water and suspended particles are uniform, \ie, $\boldsymbol{\beta}$ and $\mathbf{A}$ are consistent along a ray. Then, $I\left( \mathbf{r}_{\text{w}} \right)$'s approximation is:
\begin{equation}
\small
\begin{aligned}
   I\left( \mathbf{r}_{\text{w}} \right) \approx e^{-\boldsymbol{\beta} d}J\left( \mathbf{r}_{\text{w}} \right) + \left( {1 - e^{-\boldsymbol{\beta} d}} \right)\mathbf{A}~.
\end{aligned}
\label{eq:I_w}
\end{equation}
The simplified form of scattering rendering in \autoref{eq:I_w} is widely recognized in underwater imaging \cite{jaffe1990computer,levy2023seathru} and dehazing \cite{he2010single,nayar1999vision}, and can be discretized to:
{\small
\begin{flalign}
 I(\mathbf{r}_{\text{w}}) & \approx e^{-\boldsymbol{\beta} d}~{\sum\nolimits_{i} \! {w_{\text{o}}^{i}\mathbf{c}_{\text{o}}^{i}}} + \left( {1 - e^{-\boldsymbol{\beta} d}} \right)\mathbf{A}~,\\
 w_{\text{o}}^{i} &= \left( {1 - e_{i}^{- \sigma_{\text{o}}^{i}{({t_{i + 1} - t_{i}})}}} \right)T_{\text{o}}^{i}~,
 \label{eq:rb-}\\
 T_{\text{o}}^{i} &= {\exp\left( {\sum\nolimits_{j=0}^{i-1}{\sigma_{\text{o}}^{j}\left( {t_{j + 1} - t_{j}} \right)}} \right)}~,
\label{eq:17++}
\end{flalign}
}
where $w_{\text{o}}^{i}$ donates weight of color $\mathbf{c}_{\text{o}}^{i}$;   $\boldsymbol{\beta}, \textbf{A}\in \mathbb{R}^{3}$ in RGB space are uniform in NeuroPump; object samples' density field is $\sigma_{\text{o}}^{i}\in [0,1] $; and $d$ is the depth of pixel $\boldsymbol{x}$.

Our goal is to simultaneously obtain an implicit representation of the in-air (without water) 3D scene, \ie, \textit{in-air} object radiance field $f_\Theta:(\mathbf{x}, \mathbf{d})\to (\mathbf{c}_\text{o},\sigma_\text{o})$, the absorption and scattering factor of water and suspended particles $\boldsymbol{\beta}$, and the global background light $\mathbf{A}$. Once learned, we can render the in-air appearance of the 3D scene by:
\begin{equation}
\small
\begin{aligned}
   \hat{J}\left( \mathbf{r}_{\text{a}} \right) \approx{\sum\nolimits_{k}{w_{\text{o}}^{k}\mathbf{c}_{\text{o}}^{k}}}~,
\end{aligned}
\label{eq:J_a}
\end{equation}
where $\hat{J}\left( \mathbf{r}_{\text{a}} \right)$ is the learned in-air radiance of ray $\mathbf{r}_{\text{a}}$. Note that since the refraction and attenuation from water and suspended particles are excluded from \autoref{eq:J_a}, $w_{\text{o}}^k$ and $\mathbf{c}_\text{o}^k$ are in-air object visibility and color. The image that consists of $\hat{J}\left( \mathbf{r}_{\text{a}} \right)$ should have a geometrically and photometrically rectified appearance, as if the water in \autoref{eq:PF1} is pumped out.

\section{Implementation details}
\label{section:Implementation}
Similar to SeaThru-NeRF \cite{levy2023seathru}, we leverage the mip-NeRF 360 architecture \cite{barron2022mip} as our base model (\autoref{fig:flowchart}).
NeuroPump is trained using the reconstruction loss \cite{mildenhall2022nerf, levy2023seathru} between the camera-captured and the model inferred underwater images:
\begin{equation}
\small
\begin{aligned}
\mathcal{L}_{\text{recon}} = \left( \frac{\hat{I}(\mathbf{r}_{\text{w}}) - I(\mathbf{r}_{\text{w}})}{\text{sg}\big( \hat{I}(\mathbf{r}_{\text{w}}) \big) + \epsilon} \right)^{2}~,
\end{aligned}
\label{eq:mseloss}
\end{equation}
where $\hat{I}(\mathbf{r}_{\text{w}})$ represents the color of the sampled ray, and $\text{sg}(\cdot)$ denotes the stop-gradient function with $\epsilon = 10^{-3}$. We also introduce additional constraints on $\mathbf{A}$ and $\boldsymbol{\beta}$ below.

\noindent\textbf{Color cast loss.} Underwater images' \textit{global} color cast is mainly due to back-scatter component $A(1-T_{\text{m}}(d))$, and this prior knowledge can be applied to condition NeuroPump optimization. 
We obtain a rough global color cast ratio $\boldsymbol{\gamma}$ by averaging all image pixel values for each color channel. The color cast loss is imposed by optimizing the following loss function:
\begin{equation}
\small
\begin{split}
\mathcal{L}_{\text{cast}} &= \left| \frac{\hat{A}_{g}(1 - T_{g}(d))}{\hat{A}_{b}(1 - T_{b}(d))} - \frac{\boldsymbol{\gamma}_g}{\boldsymbol{\gamma}_b} \right| \\
&+ \sum_{c\in\{g,b\}} \left( \max \left( \frac{\hat{A}_{r}(1 - T_{r}(d))}{\hat{A}_{c}(1 - T_{c}(d))} - \frac{\boldsymbol{\gamma}_{r}}{\boldsymbol{\gamma}_{c}}, 0 \right) \right),\\
\label{eq:21Lcast}
\end{split}
\end{equation}
where $d$ is the depth of the first object point of ray $\mathbf{r}_{\text{w}}$; and $T_{c}(d)$ is:
\begin{equation}
\small
T_{c}(d) = {\exp\left({- \hat{\beta}_{c}~\mathrm{sg}(d)}\right)}~.
\label{eq:21+}
\end{equation}
This loss can consider the optimization of both back-scatter parameters and underwater unattenuated direct radiance $\hat{J}(\mathbf{r}_\text{w})$ and $\hat{J}(\mathbf{r}_\text{a})$, and its advantage is shown in \autoref{tab:ablation}.

Additionally, we employ the losses $\mathcal{L}_\text{prop}$ and $\mathcal{L}_\text{dist}$ from mip-NeRF 360 \cite{barron2022mip}.
$\mathcal{L}_\text{dist}$ minimizes the weighted distances between all pairs of interval midpoints and the weighted size of each interval. This promotes efficient density representation.
$\mathcal{L}_\text{prop}$ penalizes the difference between the distributions of object weights in the `original' and `proposed' samplings.
In summary, our training loss is:
\begin{equation}
\small
\begin{aligned}
\mathcal{L} = \mathcal{L}_\text{recon} + \mathcal{L}_\text{prop} + \lambda \left(\mathcal{L}_\text{dist} + \mathcal{L}_\text{cast}\right)~,
\end{aligned}
\label{eq:21loss}
\end{equation}
where $\lambda$ is a weight coefficient. 

\noindent\textbf{Brightness compensation.}\label{radiance intensity constraint}
Due to water attenuation, the unattenuated direct radiance of the underwater object (without the global background light) $\hat{J}(\mathbf{r}_\text{a})$ is smaller than the unattenuated direct radiance of the object in the air $J(\mathbf{r}_\text{a})$. Therefore, $\hat{J}(\mathbf{r}_\text{a})$ should be compensated to match the brightness of $J(\mathbf{r}_\text{a})$ \cite{akkaynak2019sea}. 
Inspired by Sea-thru \cite{akkaynak2019sea},
we use a global scale factor for compensation:
\begin{equation}
\small
W = \frac{\text{mean}(\hat{I}(\mathbf{r}_{\text{a}}))}{\text{mean}(\hat{J}(\mathbf{r}_{\text{a}}))}
~;~~\hat{J}(\mathbf{r}_\text{a})  \xleftarrow{} \max(1,W)\hat{J}(\mathbf{r}_{\text{a}})~,
\label{eq:22}
\end{equation}
where $\hat{I}(\mathbf{r}_{\text{a}})$ is the geometrically rectified underwater image. Note that this compensation is performed in linear space, and finally we convert the compensated images to sRGB using gamma 2.4. 

\noindent\textbf{Training details.} NeuroPump is implemented using JAX \cite{jax2018github} based on the Multi-NeRF \cite{multinerf2022} framework.
The initial learning rate undergoes log-linear annealing from $2.5\times10^{-4}$ to $2.5\times10^{-5}$ during the first 50,000 iterations.
The model is trained for 150,000 iterations with a batch size of 2,048.
The absorption and scattering parameters are optimized starting from the 35,000-th iteration. We initialize $\textbf{A}$ to [0.9, 0.9, 0.9] and bound it within [0, 1], and $\boldsymbol{\beta}$ to [0.4, 0.2, 0.2] and bound it within [0.1, 1], and 
set $\lambda$ to $10^{-3}$.

\subsection{Underwater camera pose}
\label{sec:pose_estimation}

Similar to previous NeRF-based approaches \cite{barron2022mip,mildenhall2021nerf,levy2023seathru}, our model also needs a preliminary step of camera pose estimation using COLMAP \cite{schoenberger2016sfm, schoenberger2016mvs}. 
However, the perspective camera model \cite{zhang2000flexible} in COLMAP does not account for underwater refraction, and directly running it with underwater images will cause inaccurate camera parameters estimation. Previous studies remove refraction using a dome port \cite{levy2023seathru,venkatramanan2022waternerf, Zhang_2023_beyond}. Owning to our refraction model \autoref{sec:refraction}, we can rectify refraction without using a dome port. 

As shown in \autoref{fig:geo_ref}, denote $\mathbf{x}=(x, y, z)$ as a 3D scene point, and the corresponding underwater and in-air pixel coordinates as $\boldsymbol{x}=(u,v)$ and $\boldsymbol{x}'=(u',v')$, respectively. To obtain the correct camera pose, we should use the 3D-2D correspondences between $\mathbf{x}$ and $\boldsymbol{x}'$, instead of $\boldsymbol{x}$.  Fortunately,  our refraction model enables us to establish the pixel mapping between $\boldsymbol{x}$ and $\boldsymbol{x}'$ by:
\begin{equation}
\small
\begin{aligned}
   &( u',~v') = \left( hu,~hv \right) ;~~\\
   &h = \frac{s\tan(\phi_{\text{a}}(\boldsymbol{x})) + {\left(z - s\right)}\tan(\phi_{\text{w}}(\boldsymbol{x}))}{z\tan(\phi_{\text{a}}(\boldsymbol{x}))}~,\\
\end{aligned}
\label{eq:h}
\end{equation}
where $s$ is the distance between the optical center and the lens case interface;  $z-s$ is the perpendicular distance between $\mathbf{x}$ and the camera lens case interface, as shown in \autoref{fig:geo_ref}.  The above equation holds when: (1) the distance from the optical center to the lens case interface is short, \ie, $s\approx0$, then  $h=\tan(\phi_{\text{a}})/\tan(\phi_{\text{w}})$; 
or (2) $z-s$ is uniform for all object points, \eg, when capturing a frontal view of a flat checkerboard. See Supp. for derivation. In our case, we assume that $s\approx0$ for GoPro Hero 8 camera, and apply \autoref{eq:h} to rectify underwater image for COLMAP. Note that the rectified images have a narrower field of view, and may result in black edges (\autoref{fig:flowchart}), but it is sufficient for COLMAP's keypoint matching and pose estimation.

\section{Benchmark dataset}
\label{Benchmark dataset}

To the best of our knowledge, there is no dataset with real captured in-air ground truth to estimate underwater image rectification. Previous in-air ground truth was either synthetic \cite{li2020underwater, ye2022underwater} or output of prior underwater color restoration methods \cite{li2019underwater}, while a real captured benchmark dataset with the corresponding in-air ground truth images is desired for accurate evaluation. In this paper, we propose the first real captured 360 underwater dataset with paired in-air ground truth. 

\subsection{Data collection}
\label{Data collection}
\vspace{-1mm}

A key requirement for aligned underwater and in-air images is that the global illumination, objects, camera poses, \etc, should be consistent between the underwater and the corresponding in-air scenes. 
However, it is difficult if not impossible to pump out water for paired underwater and in-air images captured in the wild. 
Therefore, following the examples of other previous works, including image dehazing \cite{ancuti2018haze, filin2023single, Chen2024dehazenerf}, robotics \cite{skinner2017automatic}, motion blur restoration \cite{kohler2012recording}, flow estimation \cite{zingg2010mav, fei2013comprehensive}, and water surface reconstruction \cite{wang2023neref, thapa2020dynamic}, we initiate our benchmark dataset using a lab environment to strictly control these factors, and captured five setups with corresponding in-air ground truth. We show our dataset's configurations in \autoref{tab:data-info}. 

\begin{table}[!t]
    \caption{Our benchmark dataset is captured by GoPro Hero 8 in linear mode. The images are downsampled from 1920 $\times$ 1080 to 960 $\times$ 540. The camera parameters and global illumination are consistent for both in-air and underwater images of each setup.} 
    
    \label{tab:data-info}    
    \centering

    \fontsize{7pt}{7pt}\selectfont
    \begin{tabular}{@{}l@{\hspace{.8mm}} c @{\hspace{.8mm}}c@{\hspace{.8mm}} l@{\hspace{1.3mm}}l@{}}

        \toprule
        Data name & Crop size & \#Views & Different global illumination & Turbidity\\
 \midrule
 Penguin hero & 955$\times$520 & 65 & Sun lamp (white) &Slight\\
 Penguin flower & 955$\times$520 & 65 & Sun lamp+Natural light (white) &Slight\\
 Lying cow & 945$\times$525 & 66 & Sun lamp+Natural light (yellow-white) &Slight\\
        Totoro & 955$\times$530 & 66 & Sun lamp+Fill light lamp (red-green)& Moderate\\
        Hamster & 960$\times$534& 66& Sun lamp+Fill light lamp (orange)& Heavy\\ 
        \bottomrule
    \end{tabular}
\end{table}

\noindent\textbf{Underwater and in-air images alignment.} 
To achieve aligned image pairs, (1) a tripod, an overhead stick, and a universal clip are utilized to stabilize the camera. The water tank is rotated on a turntable to simulate camera motion for 360 view.
(2) To ensure consistent camera poses between the underwater and the in-air image pairs, we printed a circular ruler and attached it to the turntable, so that both phases share the same marks during video recording in one environment setup.
(3) We further manually align image pairs (\ie, refraction pre-rectified in \autoref{sec:pose_estimation} and captured in-air (GT)) for more accurate estimation.
(4) Tap water contains soluble gas that may lead to dense bubbles on objects, creating disparities between the underwater and the in-air images. Therefore, we agitated the water to release gas and reduce bubble production. 

\noindent\textbf{COLMAP requirements.} Specular reflections and a limited number of keypoints will hinder COLMAP pose estimation. Therefore, we use frosted inner tank and feature-rich objects. Moreover, highly turbid water will also cause failed keypoints extraction, and this challenge is pervasive across all NeRF-based methods that use COLMAP. Therefore, we limited the maximum turbidity in our environment setups to avoid COLMAP failure. Additional data collection details can be found in the Supp..

\begin{figure*}[!t]
  \centering

    \includegraphics[width=1.0\linewidth]{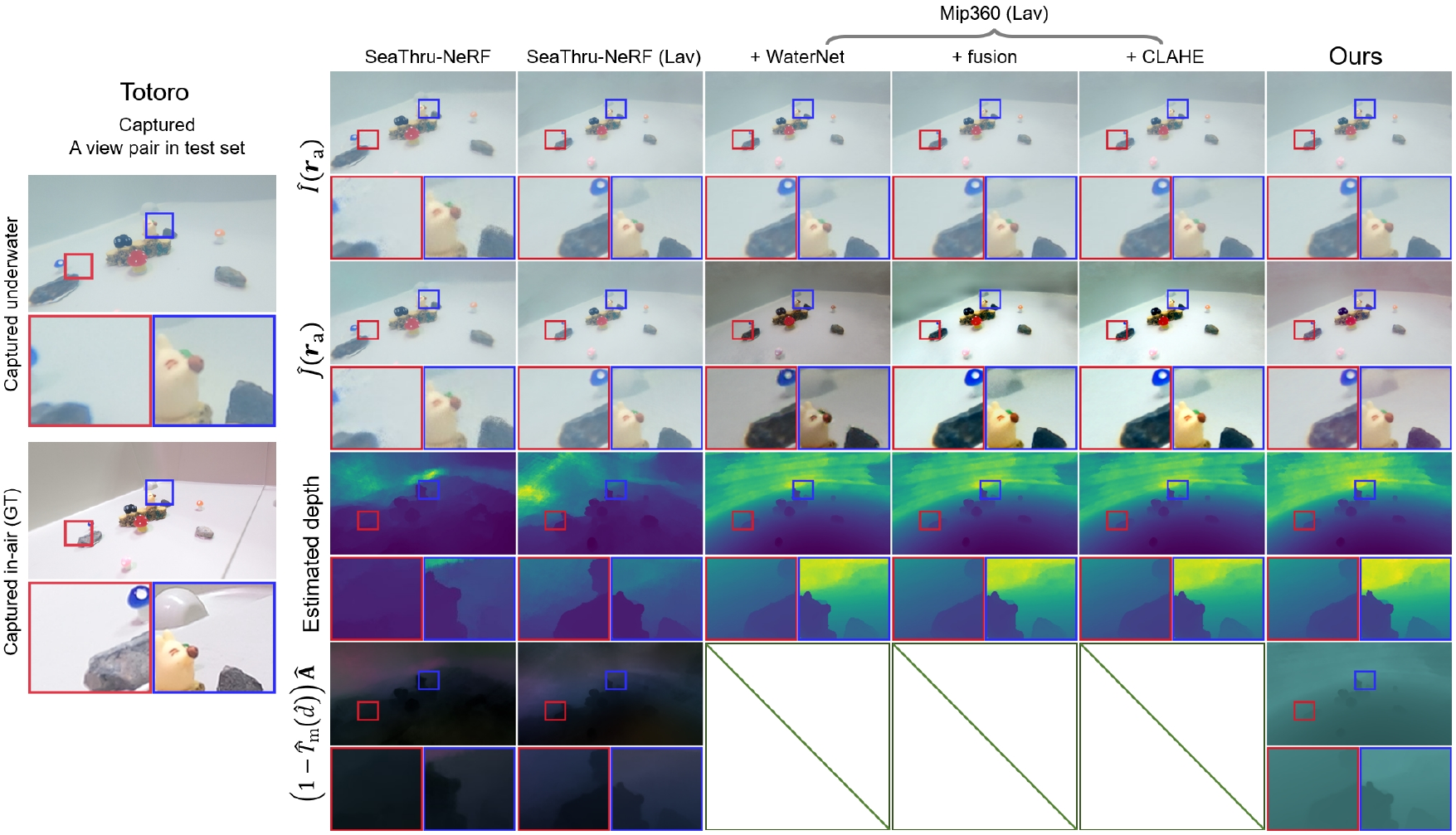}
    
  \caption{Rectified underwater images and intermediate results. The 1st row is the underwater image with \textit{geometric rectification only}. Clearly, the original SeaThru-NeRF's result looks like a zoomed-in version of the in-air ground truth, because it cannot rectify refraction. The 2nd row is the underwater image with \textit{joint geometric and color rectification (the final target)}. Note that  
  SeaThru-NeRF and SeaThru-NeRF (Lav) cannot accurately rectify underwater color, and the estimated depth (the 3rd row) and back-scatter (the 4th row) are also inferior. Mip360 (Lav)-based methods show differences in brightness and saturation compared to the in-air ground truth. In comparison, our NeuroPump shows clear advantages in all results. See more setups and results in Supp.. }
   \label{fig:airwater}
\end{figure*}

\vspace{-1mm}
\section{Experiments}
\label{experiment results}
\vspace{-1mm}

\begin{table}[!t]
    \centering
    \setlength{\tabcolsep}{2pt}
    \caption{Experiment settings. Model settings on pose estimation, camera intrinsics calibration, and refraction removal method. For pose, linear means the poses are estimated by COLMAP using underwater images, without modeling refraction. For camera intrinsics calibration, in-air and underwater indicate where the OpenCV calibration chessboard images were taken. Others refers to underwater single image color rectification methods: WaterNet \cite{li2019underwater}, fusion \cite{ancuti2017color} and CLAHE \cite{zuiderveld1994contrast}.}
    \label{tab:setting}
    \vspace{-2mm}
    \fontsize{8pt}{8pt}\selectfont
    \begin{tabular}{l@{\hspace{2mm}} l@{\hspace{2mm}} l@{\hspace{2mm}} l}
    \toprule
        Model & Pose & Intrin. Calib. & Refraction removal \\ 
    \midrule
        SeaThru-NeRF \cite{levy2023seathru} & Linear &  In-air & As linear (ignore) \\ 
        SeaThru-NeRF (Lav) & Lavest & Underwater & Lavest approx. \cite{lavest2000underwater} \\
        Mip360 \cite{barron2022mip}  (Lav) + Others & Lavest & Underwater & Lavest approx. \cite{lavest2000underwater}\\ 
        Ours & Ours & In-air & Snell's law \cite{hecht2002optics} \\
    \bottomrule
    \end{tabular}
\end{table}

\begin{table}   
    \centering
    \caption{Quantitative comparisons on joint geometric and color rectification (\ie, 
 obtaining the in-air image $\hat{J}(\mathbf{r}_{\text{a}})$, the final target) and geometric rectification  (\ie, 
 only removing refraction $\hat{I}(\mathbf{r}_{\text{a}})$)
. Note that the metrics of $\hat{J}(\mathbf{r}_\text{a})$ are worse than $\hat{I}(\mathbf{r}_\text{a})$ for the Mip360 (Lav) + Others baselines, because the color rectification methods focus on improving color contrast, saturation and brightness that may be different from the real in-air GT images. The results are averaged on the testing set of the five setups. In the last column, the same value of Ours and Mip360 (Lav) + Others baselines is a coincidence due to average.     \label{tab:results}
See Supp. for individual comparison.}

 \vspace{-1.5mm}
    \setlength{\tabcolsep}{1pt}
    \small
    \setlength{\tabcolsep}{54pt}
    \resizebox{\linewidth}{!}{
    \begin{tabular}{@{\hspace{0mm}}l@{\hspace{3.8mm}}l@{\hspace{2.8mm}}l@{\hspace{2.8mm}}l@{\hspace{5mm}}l@{\hspace{2.8mm}}l@{\hspace{2.8mm}}l@{\hspace{2mm}}}
    \toprule
         \multirow{2}[2]{*}{\centering Model}  & \multicolumn{3}{@{}l@{\hspace{1mm}}}{\hspace{0.5mm}$\hat{J}(\textbf{r}_\text{a})$ (geometry \& color)}  &  \multicolumn{3}{@{}l@{\hspace{1mm}}}{\hspace{2.5mm}$\hat{I}(\textbf{r}_\text{a})$ (geometry only)} \\ 
         \cmidrule(l{0mm}r{5mm}){2-4} \cmidrule(l{0mm}r){5-7}
         & PSNR$\uparrow$ & SSIM$\uparrow$ & RMSE$\downarrow$ & PSNR$\uparrow$ & SSIM$\uparrow$ & RMSE$\downarrow$ \\ \midrule 
        Seathru-NeRF \cite{levy2023seathru} & 17.1145& 0.8030& 0.1398& 17.2507& 0.8072& 0.1376\\ 
        Seathru-NeRF (Lav \cite{lavest2000underwater}) & 20.8755& 0.8966& 0.0919& 20.8999& 0.8978& 0.0917\\ 
        Mip360 \cite{barron2022mip} (Lav) + WaterNet \cite{li2019underwater} & 15.1490& 0.8443& 0.1824& \textbf{21.0935}& \textbf{0.9012}& \textbf{0.0900}\\ 
        Mip360 \cite{barron2022mip} (Lav) + fusion \cite{ancuti2017color}  & 18.7843& 0.8586& 0.1163& \textbf{21.0935}& \textbf{0.9012}& \textbf{0.0900}\\ 
        Mip360 \cite{barron2022mip} (Lav) + CLAHE \cite{zuiderveld1994contrast} & 18.8043& 0.8139& 0.1166& \textbf{21.0935}& \textbf{0.9012}& \textbf{0.0900}\\ 
        Ours & \textbf{22.7571}& \textbf{0.9008}& \textbf{0.0746}& 21.0906& 0.9009& \textbf{0.0900}\\ \bottomrule
    \end{tabular}
    }
\end{table}

\noindent\textbf{Baselines.} We compare our NeuroPump with two state-of-the-art methods, SeaThru-NeRF \cite{levy2023seathru}, mip-NeRF 360 \cite{barron2022mip} and three underwater single image color rectification methods \cite{li2019underwater, ancuti2012enhancing, zuiderveld1994contrast}. Note that each method cannot be directly applied to underwater images that are captured without a dome port if joint rectification of geometry and color is required. For fair comparisons, we combine Lavest's refraction removal \cite{lavest2000underwater} for SeaThru-NeRF and mip-NeRF 360, donated as SeaThru-NeRF (Lav) and Mip360 (Lav). Additionally, Mip360 (Lav) is further combined with three color rectification methods \cite{li2019underwater, ancuti2012enhancing, zuiderveld1994contrast}, denoted as Mip360 (Lav) + Others.
It is worth noting that Lavest \etal \cite{lavest2000underwater} approximates underwater refraction using camera lens radial distortion, particularly for cameras without a lens case interface. This method may be less accurate when camera lens case significantly deviates from the optical center \cite{jordt2012refractive, chadebecq2017refractive}, and the refractive indices are entangled in the model. While our method leverages Snell's law \cite{hecht2002optics}, and can achieve more accurate refraction modeling across various camera models. Moreover, our approach can also be applied to simulated novel views with new optical parameters.
The detailed experiment settings are shown in \autoref{tab:setting}.

For \textbf{Underwater image rectification},  
we compared our NeuroPump with all baselines on geometric rectification and simultaneous geometric and color rectification tasks. Denote the geometric rectified image as $\hat{I}(\mathbf{r}_{\text{a}})$ (\ie, color is not rectified), and in-air image as $\hat{J}(\mathbf{r}_{\text{a}})$ (\ie, both geometry and color are rectified). 
The quantitative comparisons are shown in \autoref{tab:results}. 

\begin{figure}[!t]
  \centering
    \includegraphics[width=0.995\linewidth]{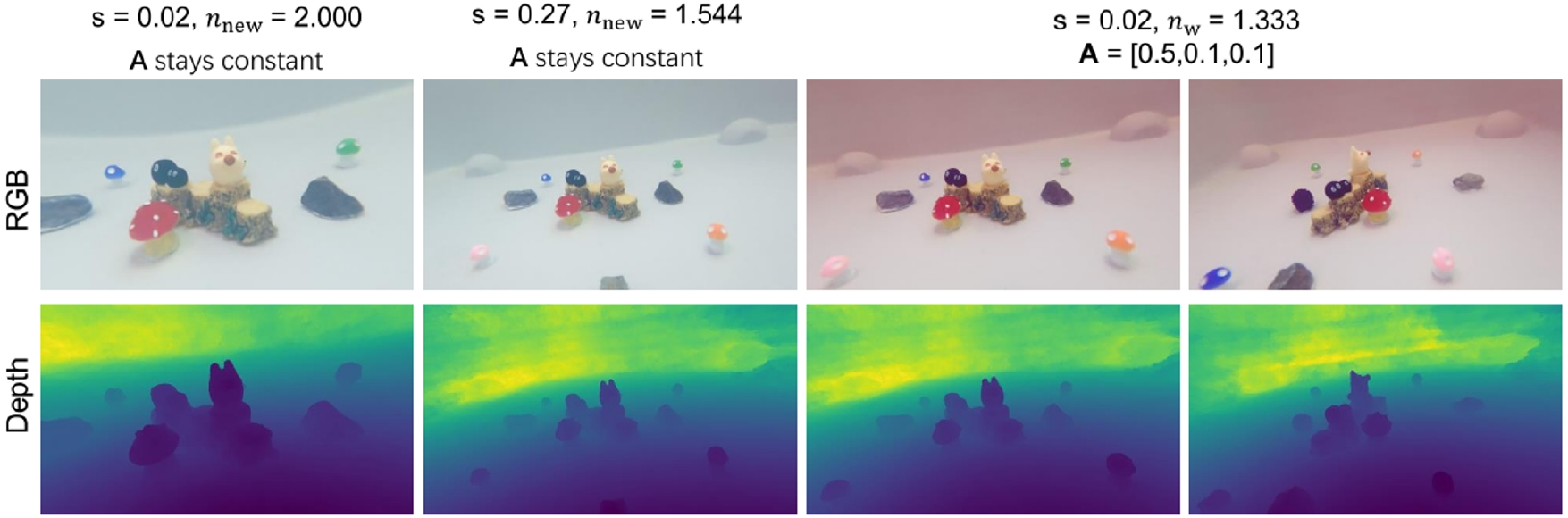}
   \caption{Novel view and optical parameter synthesis. The camera pose and the perpendicular distance from optical center to the lens case interface $s$, the medium refractive index $n_{\text{new}}$, and global background light $\textbf{A}$ are varied for new image synthesis.}
   \label{fig:synthesis}
\end{figure}

For \textbf{geometric rectification} $\hat{I}(\mathbf{r}_{\text{a}})$, our model is almost on par with Mip360 (Lav) $\hat{I}(\mathbf{r}_{\text{a}})$ on PSNR/RMSE/SSIM, because GoPro Hero 8 camera has a very small optical center to lens case distance $s$ (\autoref{fig:geo_ref}), and Lavest's refraction removal method \cite{lavest2000underwater} works well in this case. But in our method, the optical parameters, such as refractive index, $s$ and global background light are decoupled, and we can synthesize novel views with new optical parameters (\autoref{fig:synthesis}). 

For \textbf{joint geometric and color rectification (the final target)} $\hat{J}(\mathbf{r}_{\text{a}})$, as shown in \autoref{fig:airwater} and \autoref{tab:results}, our NeuroPump outperforms the other baselines in terms of visual appearance (closer to the captured in-air ground truth) and higher PSNR/SSIM and lower RMSE. 
\textbf{Notably,} this indicates simultaneous color and geometric distortions rectification (Ours) works better than rectifying them separately (Mip360 (Lav) + Others).

\begin{table} [!t] 
    \centering
    \caption{Ablation studies. The results are averaged on the testing set of the five setups.}
    \label{tab:ablation}
    \vspace{-1mm}
    \fontsize{8.5pt}{9pt}\selectfont
    \begin{tabular}{@{\hspace{1.5mm}}l@{\hspace{2.mm}} @{\hspace{1.5mm}}l@{\hspace{1.5mm}} @{\hspace{1.5mm}}l@{\hspace{1.5mm}} @{\hspace{1.5mm}}l@{\hspace{1.5mm}}}
        \toprule
        Model & PSNR$\uparrow$ & SSIM$\uparrow$ & RMSE$\downarrow$  \\
        \midrule
        Ours w/o brightness compensation & 21.3880& 0.8983& 0.0901\\
        Ours w/o $\mathcal{L}_{\text{cast}}$ & 21.7346& \textbf{0.9010}& 0.0830\\
       Ours full (NeuroPump)& \textbf{22.7571}& 0.9008& \textbf{0.0746}\\
        \bottomrule
    \end{tabular}
\end{table}

\subsection{Ablation study}
\vspace{-1mm}
To show the effectiveness of our model design, we performed ablation studies using three different versions of our NeuroPump: (a) NeuroPump without brightness compensation, (b) NeuroPump without $\mathcal{L}_{\text{cast}}$, and (c) NeuroPump. The quantitative results in \autoref{tab:ablation} show that the SSIM is similar for all models, while the PSNR and RMSE of degraded NeuroPump models are clearly lower, demonstrating the effectiveness of both the brightness compensation and $\mathcal{L}_{\text{cast}}$ in our model training and design. 

\subsection{Novel view and optical parameter synthesis}
\label{further rendering}
\vspace{-1mm}
Snell's law \cite{hecht2002optics} plays a pivotal role in our model, offering clarity, decoupling, and applicability across various camera models. It empowers the trained model to generate underwater views with novel optical parameters, resulting in captivating effects. Examples of these novel views with new optical parameters, are visually depicted in \autoref{fig:synthesis} and our project page: \href{https://ygswu.github.io/NeuroPump.github.io/}{https://ygswu.github.io/NeuroPump.github.io/}. Clearly, the synthesized images look geometrically and photometrically realistic.

\section{Conclusion and limitations}
\vspace{-2mm}
Previous studies focus on rectifying either the color or the geometry, and in practice it may be cumbersome to address the two tasks one-by-one. 
We present a the first model that simultaneously restores both geometric and color distortions for underwater images. By explicitly decoupling the optical parameters, our model can simulate novel views with new optical parameters of underwater scenes. Our work also addressed the lack of real paired ground truth images by obtaining a real captured underwater 360 benchmark dataset with paired images. 

\noindent\textbf{Limitations and future work.} The assumption that the distance between the camera lens case interface and the optical center may not apply to all cameras. In addition, our dataset has limited scene depth and does not include highly turbid water, because some NeRF-based approaches (including ours) that rely on COLMAP may fail in highly turbid waters due to lack of feature points. Simultaneously estimating the camera intrinsics, extrinsics and water parameters may be a direction to address this limitation. Future work will also focus on exploring underwater pose estimation using fewer feature points, comparing methods on different cameras, more complex water and light conditions (\textit{e.g.}, challenging natural water) and formulating forward-scattering for highly turbid water.

{
    \small
    \bibliographystyle{ieeenat_fullname}
    \bibliography{main}
}

\clearpage
\setcounter{page}{1}
\maketitlesupplementary
\appendix

\maketitle

\section{Overview}
The supplementary material first provides in-depth derivations for the main paper's \autoref{sec:problem_formulation} and \autoref{sec:pose_estimation}, and complete experiment results on our benchmark dataset. In addition, we include synthesized \textbf{videos} of novel views and new optical parameters on our project page: \href{https://ygswu.github.io/NeuroPump.github.io/}{https://ygswu.github.io/NeuroPump.github.io/}.

\section{Refraction}
\label{suppl:refraction}

Derivation of main paper's \autoref{eq:16++}, \ie:
\begin{equation}
\mathbf{d}_{\text{w}}(\boldsymbol{x}) = \textbf{Refract}\left( {\mathbf{d}_{\text{a}}(\boldsymbol{x}),\phi_{\text{a}}(\boldsymbol{x}),\phi_{\text{w}}(\boldsymbol{x})} \right). 
\end{equation}
It can be further elucidated using Rodrigues’ rotation formula:
\begin{equation}
\begin{aligned}
     \mathbf{d}_{\text{w}}(\boldsymbol{x}) &= \mathbf{d}_{\text{a}}(\boldsymbol{x}){\cos\left( {\phi_{\text{a}}(\boldsymbol{x}) - \phi_{\text{w}}(\boldsymbol{x})} \right)} \\
  &\quad + \left( {\mathbf{u}(\boldsymbol{x}) \times \mathbf{d}_{\text{a}}}(\boldsymbol{x}) \right){\sin\left( {\phi_{\text{a}}(\boldsymbol{x}) - \phi_{\text{w}}(\boldsymbol{x})} \right)} \\
  &\quad + \left( {\mathbf{u}(\boldsymbol{x}) \cdot \mathbf{d}_{\text{a}}(\boldsymbol{x})} \right)\mathbf{u}(\boldsymbol{x})\left( {1 - {\cos\left( {\phi_{\text{a}}(\boldsymbol{x}) - \phi_{\text{w}}(\boldsymbol{x})} \right)}} \right).
\end{aligned}
\end{equation}
The rotation axis $\mathbf{u}(\boldsymbol{x})$ is given by:

\begin{equation}
\begin{aligned}
    \mathbf{u}(\boldsymbol{x}) = \mathbf{d}_{\text{a}}(\boldsymbol{x}) \times \mathbf{g}, 
\end{aligned}
\end{equation}
where $\mathbf{g}$ is the camera lens case's normal vector. 

\section{Single image geometry rectification}
\label{single_image_geometry_restoration}
\begin{figure}[t]
    \centering
        \includegraphics[width=1.0\linewidth]{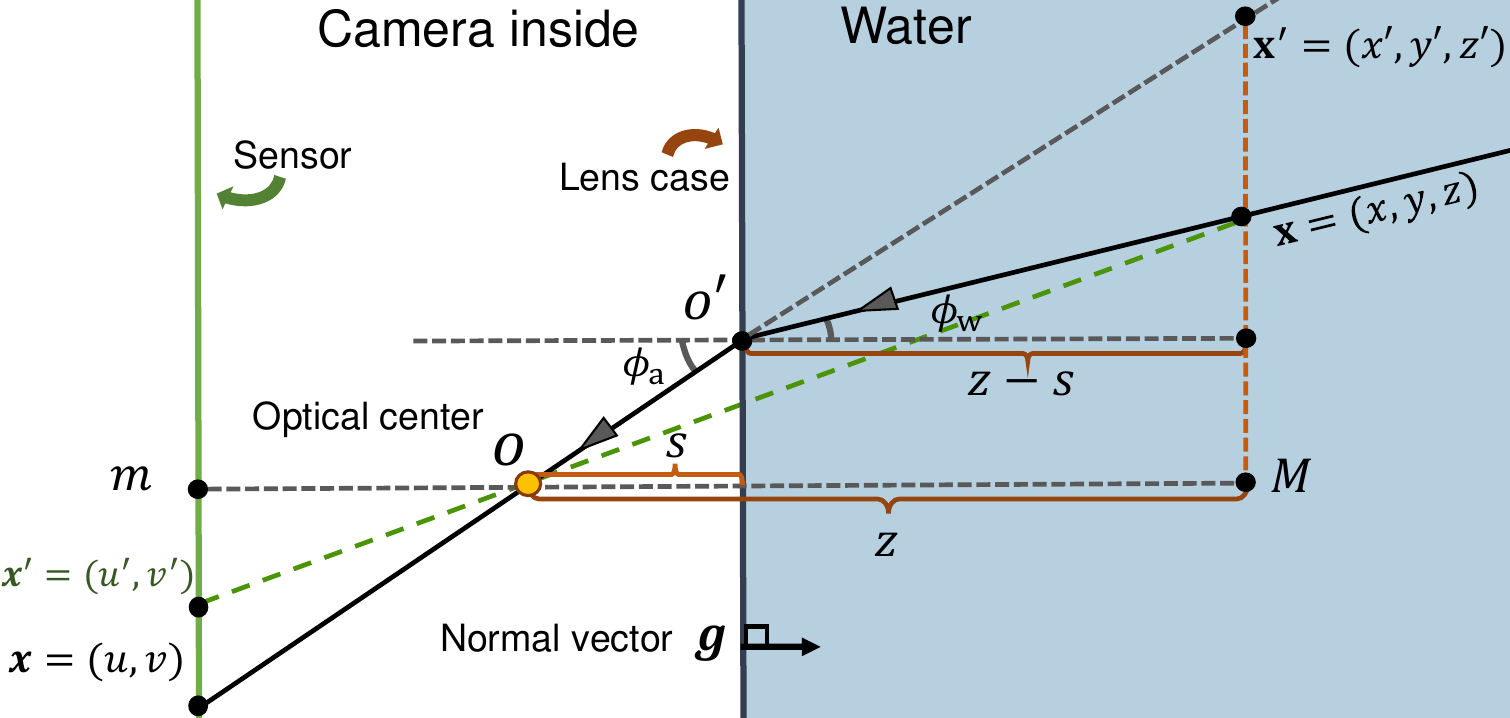}
    \caption{Underwater imaging. We assume the world origin is at the camera optical center $O=(0,0,0)$, and the camera sensor's principle point is $m=(0,0)$. Giving an object point $\mathbf{x}=(x, y, z)$ in the world space, its corresponding pixel coordinate is $\boldsymbol{x}=(u,v)$ when imaged underwater due to refraction. While $\mathbf{x}$'s actual pixel coordinate should be $\boldsymbol{x}'=(u',v')$. Moreover, $\mathbf{g}$ is the normal vector of lens case interface.}
    \label{fig:suppl_geo}
\end{figure}

Derivation of the geometric rectification factor $h$ (\ie, main paper's \autoref{eq:h}). As shown in \autoref{fig:suppl_geo}, we assume the world origin is at the camera optical center $O=(0,0,0)$, and the camera sensor's principle point is $m=(0,0)$. For an object point $\mathbf{x}=(x, y, z)$ in the world space, its corresponding pixel coordinate is $\boldsymbol{x}=(u,v)$ when imaged underwater due to refraction. However, its actual pixel coordinate should be $\boldsymbol{x}'=(u',v')$, and the relationship between $\boldsymbol{x}'$ and $\boldsymbol{x}$ is given by:
\begin{equation}
\begin{aligned}
    (u', v') &= (hu, hv)~,\\
    h &= \frac{\|\boldsymbol{x'}m\|_2}{\|\boldsymbol{x}m\|_2} = \frac{\|\mathbf{x}M\|_2}{\|\mathbf{x'}M\|_2} \\
    &= \frac{s\tan(\phi_{\text{a}}(\boldsymbol{x})) + {\left(z - s\right)}\tan(\phi_{\text{w}}(\boldsymbol{x}))}{z\tan(\phi_{\text{a}}(\boldsymbol{x}))}~.
\end{aligned}
\label{eq:h}
\end{equation}

In practice, $z$ and $s$ are unknown, and we can solve for $h$ in three different cases:

\noindent(1) When $s$ is small, \eg,  $s \approx 0$, $h$ can be simplified as follows:
\begin{equation}
\begin{aligned}
   h = \frac{\tan(\phi_{\text{w}}(\boldsymbol{x}))}{\tan(\phi_{\text{a}}(\boldsymbol{x}))}~.
\end{aligned}
\label{}
\end{equation}
\noindent(2) When $z$ is uniform across object points, \eg, all object points lie on the same plane that is parallel to the camera lens sensor and lens case. Once $s/z$ is obtained, $h$ becomes:
\begin{equation}
\begin{aligned}
   h = \frac{1 + (s/z)(\tan(\phi_{\text{a}}(\boldsymbol{x})) - {\tan(\phi_{\text{w}}(\boldsymbol{x})))}}{\tan(\phi_{\text{a}}(\boldsymbol{x}))}~.
\end{aligned}
\label{eq:sup_h}
\end{equation}
\noindent(3) When $z$ varies across object points, each object point's depth is needed to solve for accurate $h$.

\section{Experiments}

\begin{table*}[h]  
    \centering
    \caption{Complete quantitative comparison on joint geometric and color rectification (\ie, 
 obtaining the in-air image $\hat{J}(\mathbf{r}_{\text{a}})$, the final target) and geometric rectification  (\ie, 
 only removing refraction $\hat{I}(\mathbf{r}_{\text{a}})$)
 on our benchmark dataset. Note that the metrics of $\hat{J}(\textbf{r}_\text{a})$ is worse than $\hat{I}(\textbf{r}_\text{a})$ for the Mip360 (Lav) + Others baselines, because the color rectification methods focus on improving color contrast, saturation and brightness that may be different from the real in-air GT images. }
    \fontsize{10.5pt}{11.5pt}\selectfont
    \setlength{\tabcolsep}{1pt}
    \label{tab:breakdown}
    \setlength{\tabcolsep}{54pt}

    \begin{tabular}{@{\hspace{0mm}}l@{\hspace{5.4mm}}l@{\hspace{5.2mm}}l@{\hspace{2.mm}}l@{\hspace{2.mm}}l@{\hspace{2mm}}l@{\hspace{1.mm}}l@{\hspace{1.mm}}l@{\hspace{0mm}}}
    \toprule
         \multirow{2}[2]{*}{\centering Data} & \multirow{2}[2]{*}{\centering Model}  & \multicolumn{3}{@{}l@{\hspace{1mm}}}{\hspace{0.5mm}$\hat{J}(\textbf{r}_\text{a})$ (geometry \& color)}  &  \multicolumn{3}{@{}l@{\hspace{1mm}}}{\hspace{2.5mm}$\hat{I}(\textbf{r}_\text{a})$ (geometry only)} \\ 
         \cmidrule(l{0mm}r{2mm}){3-5} \cmidrule(l{0mm}r){6-8}
         & &PSNR$\uparrow$ & SSIM$\uparrow$ & RMSE$\downarrow$ & PSNR$\uparrow$ & SSIM$\uparrow$ & RMSE$\downarrow$ \\ \midrule 
        \multirow{6}{*}{\centering Penguin hero} & Seathru-NeRF \cite{levy2023seathru} & 16.8309 & 0.7847 & 0.1446 & 16.9565 & 0.7891 & 0.1426 
\\ 
        & Seathru-NeRF (Lav \cite{lavest2000underwater}) & 21.2786 & 0.8850 & 0.0865 & 21.0708 & 0.8870 & 0.0885 \\ 
        & Mip360 \cite{barron2022mip} (Lav) + WaterNet \cite{li2019underwater} & 15.0396& 0.8417& 0.1806
& 21.4962 & 0.8969 & 0.0842 \\ 
        & Mip360 (Lav) + fusion \cite{ancuti2017color}  & 19.9828& 0.8655& 0.1003& 21.4962 & 0.8969 & 0.0842 \\ 
        & Mip360 (Lav) + CLAHE \cite{zuiderveld1994contrast} & 20.3073& 0.8132& 0.0965& 21.4962 & 0.8969 & 0.0842 \\ 
        & Ours & \textbf{23.0724} & \textbf{0.8965} & \textbf{0.0702} & \textbf{21.5367} & \textbf{0.8973} & \textbf{0.0838} \\ \midrule

        \multirow{6}{*}{\centering Penguin flower} & Seathru-NeRF \cite{levy2023seathru} & 16.6865 & 0.7820 & 0.1467 & 16.8767 & 0.7859 & 0.1435 
\\
        & Seathru-NeRF (Lav \cite{lavest2000underwater}) & 20.6086 & 0.8867 & 0.0934 & 20.9129 & 0.8894 & 0.0903 \\ 
        & Mip360 \cite{barron2022mip} (Lav) + WaterNet \cite{li2019underwater} & 17.7489& 0.8557& 0.1311
& \textbf{21.2685} & 0.\textbf{8930} & \textbf{0.0865} \\
        & Mip360 (Lav) + fusion \cite{ancuti2017color}  & 19.5272& 0.8556& 0.1057& \textbf{21.2685} & 0.\textbf{8930} & \textbf{0.0865} \\ 
        & Mip360 (Lav) + CLAHE \cite{zuiderveld1994contrast} & 20.3904& 0.8246& 0.0957& \textbf{21.2685} & 0.\textbf{8930} & \textbf{0.0865} \\ 
        & Ours & \textbf{22.7369} & \textbf{0.8934} & \textbf{0.0732} & 21.2379 & 0.8921 & 0.0868 \\ \midrule

        \multirow{6}{*}{\centering Lying cow} & Seathru-NeRF \cite{levy2023seathru} & 17.6290 & 0.8115 & 0.1320 & 17.6701 & 0.8151 & 0.1314 \\ 
        & Seathru-NeRF (Lav \cite{lavest2000underwater}) & 23.4194 & 0.9201 & 0.0675 & 23.6036 & 0.9211 & 0.0661 \\ 
        & Mip360 \cite{barron2022mip} (Lav) + WaterNet \cite{li2019underwater} &15.4484 &0.8546 &0.1768 & \textbf{23.8416} & \textbf{0.9242} & \textbf{0.0643} \\ 
        & Mip360 (Lav) + fusion \cite{ancuti2017color}  & 19.0020
& 0.8770
& 0.1123
& \textbf{23.8416} & \textbf{0.9242} & \textbf{0.0643} \\ 
        & Mip360 (Lav) + CLAHE \cite{zuiderveld1994contrast} & 17.0293& 0.7615& 0.1410& \textbf{23.8416} & \textbf{0.9242} & \textbf{0.0643} \\ 
        & Ours & \textbf{25.7708} & \textbf{0.9214} & \textbf{0.0516} & 23.8387 & 0.9237 & \textbf{0.0643} \\ \midrule

        \multirow{6}{*}{\centering Totoro} & Seathru-NeRF \cite{levy2023seathru} & 17.3704 & 0.8236 & 0.1355 & 17.3759 & 0.8264 & 0.1355 
\\ 
        & Seathru-NeRF (Lav \cite{lavest2000underwater}) & 20.4499 & 0.9101 & 0.0951 & 20.1975 & 0.9098 & 0.0979 \\ 
        & Mip360 \cite{barron2022mip} (Lav) + WaterNet \cite{li2019underwater} & 12.2523
& 0.8381
& 0.2479
& 20.2946 & 0.9106 & 0.0968 \\ 
        & Mip360 (Lav) + fusion \cite{ancuti2017color}  & 18.5969& 0.8730
& 0.1187
& 20.2946 & 0.9106 & 0.0968 \\ 
        & Mip360 (Lav) + CLAHE \cite{zuiderveld1994contrast} & 19.0212
& 0.8415
& 0.1119
& 20.2946 & 0.9106 & 0.0968 \\ 
        & Ours & \textbf{22.4898} & \textbf{0.9113} & \textbf{0.0751} & \textbf{20.3417} & \textbf{0.9117} & \textbf{0.0962} \\ \midrule

        \multirow{6}{*}{\centering Hamster} & Seathru-NeRF \cite{levy2023seathru} & 17.0561& 0.8135& 0.1403
& 17.3745& 0.8196& 0.1353
\\ 
        & Seathru-NeRF (Lav \cite{lavest2000underwater}) & 18.6212& 0.8814& 0.1172
& \textbf{18.7150}& \textbf{0.8820}& \textbf{0.1160}
\\ 
        & Mip360 \cite{barron2022mip} (Lav) + WaterNet \cite{li2019underwater} & 15.2562& 0.8317& 0.1757
& 18.5288& 0.8810& 0.1185
\\ 
        & Mip360 (Lav) + fusion \cite{ancuti2017color}  & 16.8127& 0.8219& 0.1450
& 18.5288& 0.8810& 0.1185
\\ 
        & Mip360 (Lav) + CLAHE \cite{zuiderveld1994contrast} & 17.2735& 0.8291& 0.1380
&  18.5288& 0.8810& 0.1185
\\ 
        & Ours & \textbf{19.7682}& \textbf{0.8816}& \textbf{0.1028}
& 18.5179& 0.8800& 0.1186
\\ \midrule
    \end{tabular}
\end{table*}

We provide the complete comparison between our NeuroPump and other baselines on our benchmark dataset in \autoref{fig:suppl_data} (excluding Totoro, as it is already included in the main paper's \autoref{fig:airwater}) and \autoref{tab:breakdown}. 
air
As shown in the \textbf{videos} of novel view synthesis, our NeuroPump's final rectification results clearly outperform other baselines both quantitatively and qualitatively. Moreover, our NeuroPump can perform new optical parameter synthesis because it explicitly decouple and learn these optical parameter.

\section{More dataset acquisition details}
\label{suppl dataset acquisition}
In addition to data acquisition details in main paper \autoref{Data collection}, camera pose consistency during in-air and underwater capturing is also worth noting. Therefore, instead of moving the camera directly, we mount the camera onto a tripod with an overhead boom, and only rotate a turntable beneath the water tank, and control the camera remotely over Wi-Fi.

\begin{figure*}[t]
    \centering
    \includegraphics[width=1.0\textwidth]{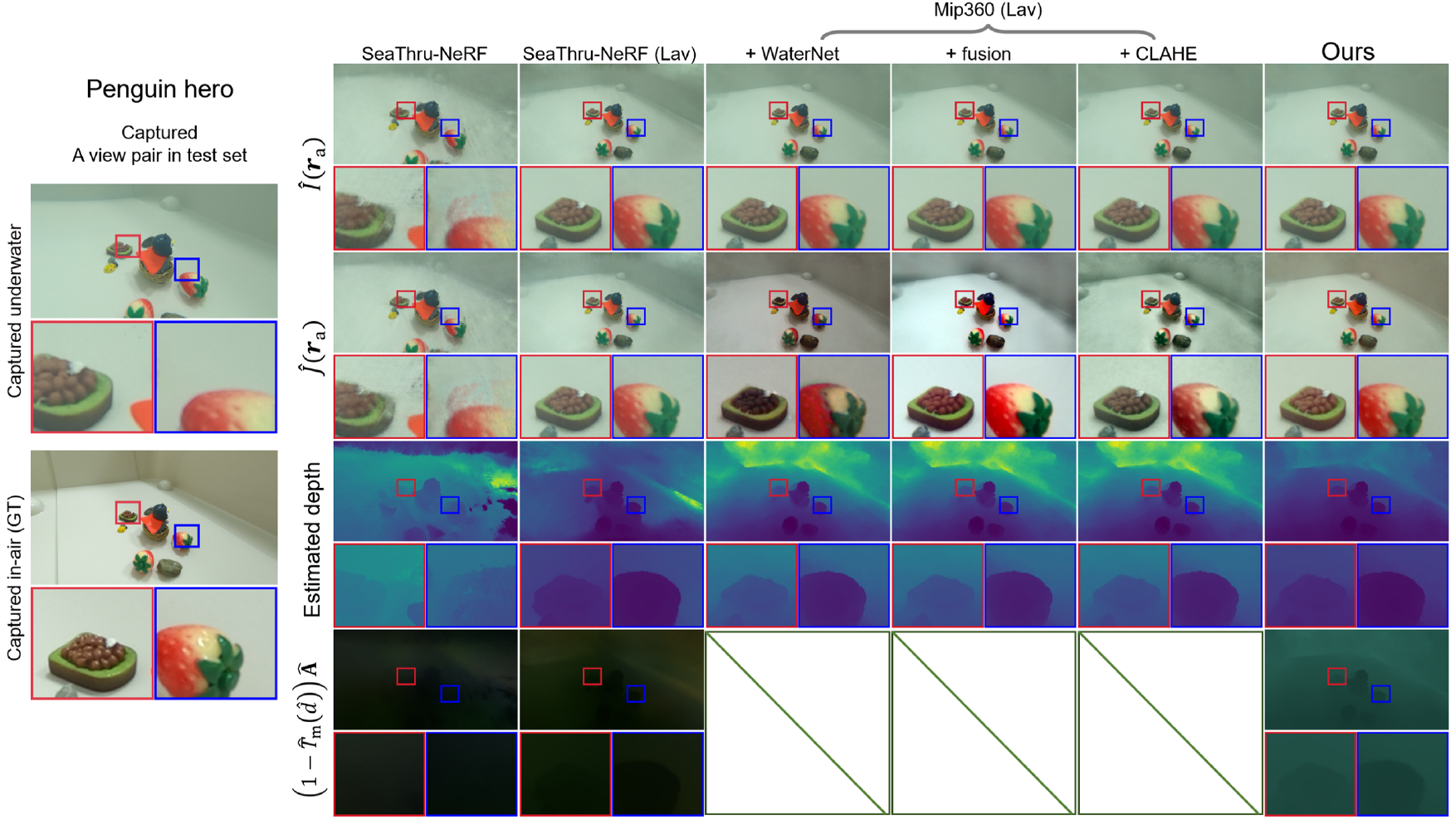}
    \centering{(a) Penguin hero}
    \includegraphics[width=1.0\textwidth]{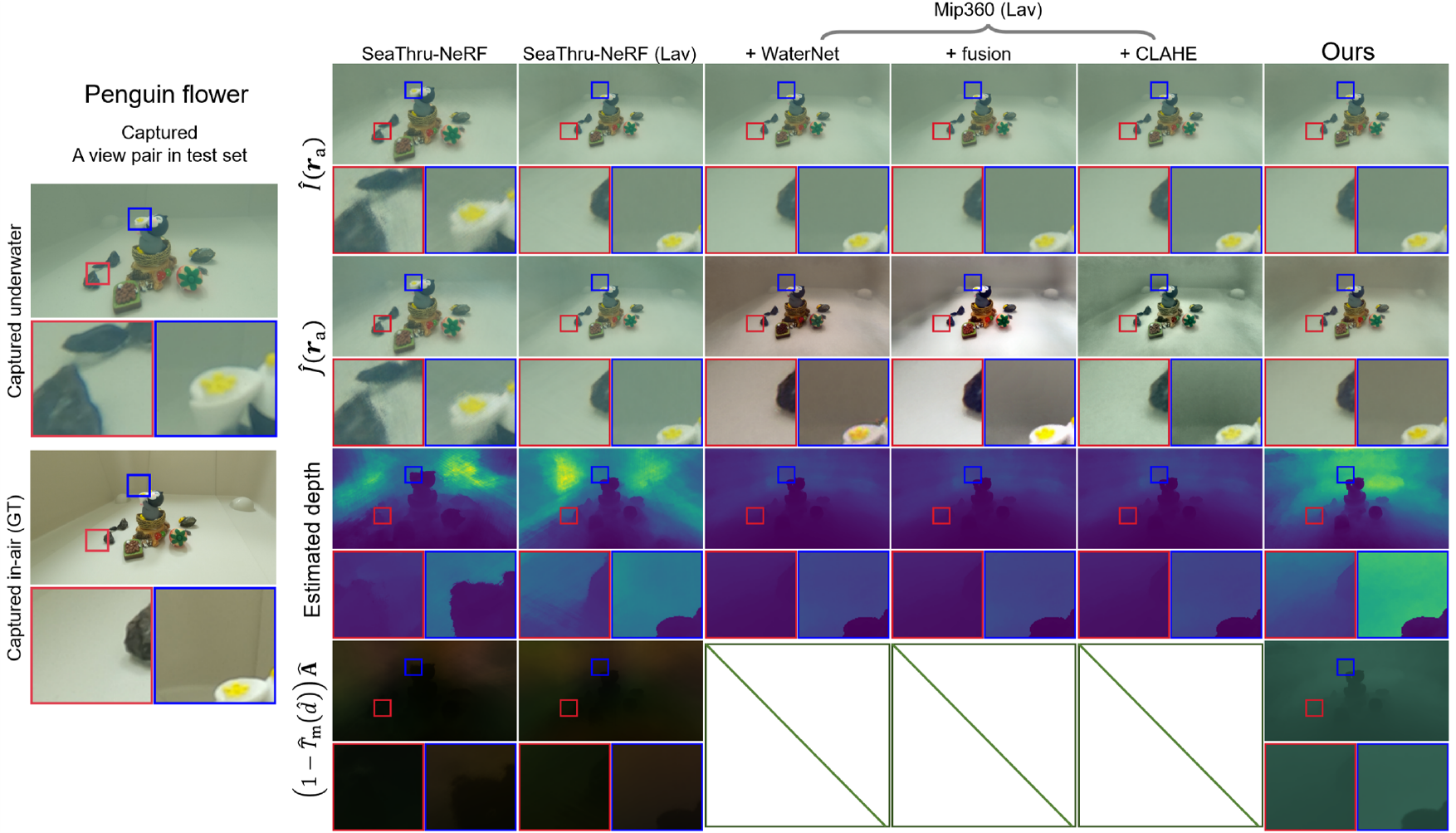}
    \centering{(b) Penguin flower}
    \label{fig:suppl_data}
\end{figure*}

\begin{figure*}[t]
        \centering
        \includegraphics[width=1.0\textwidth]{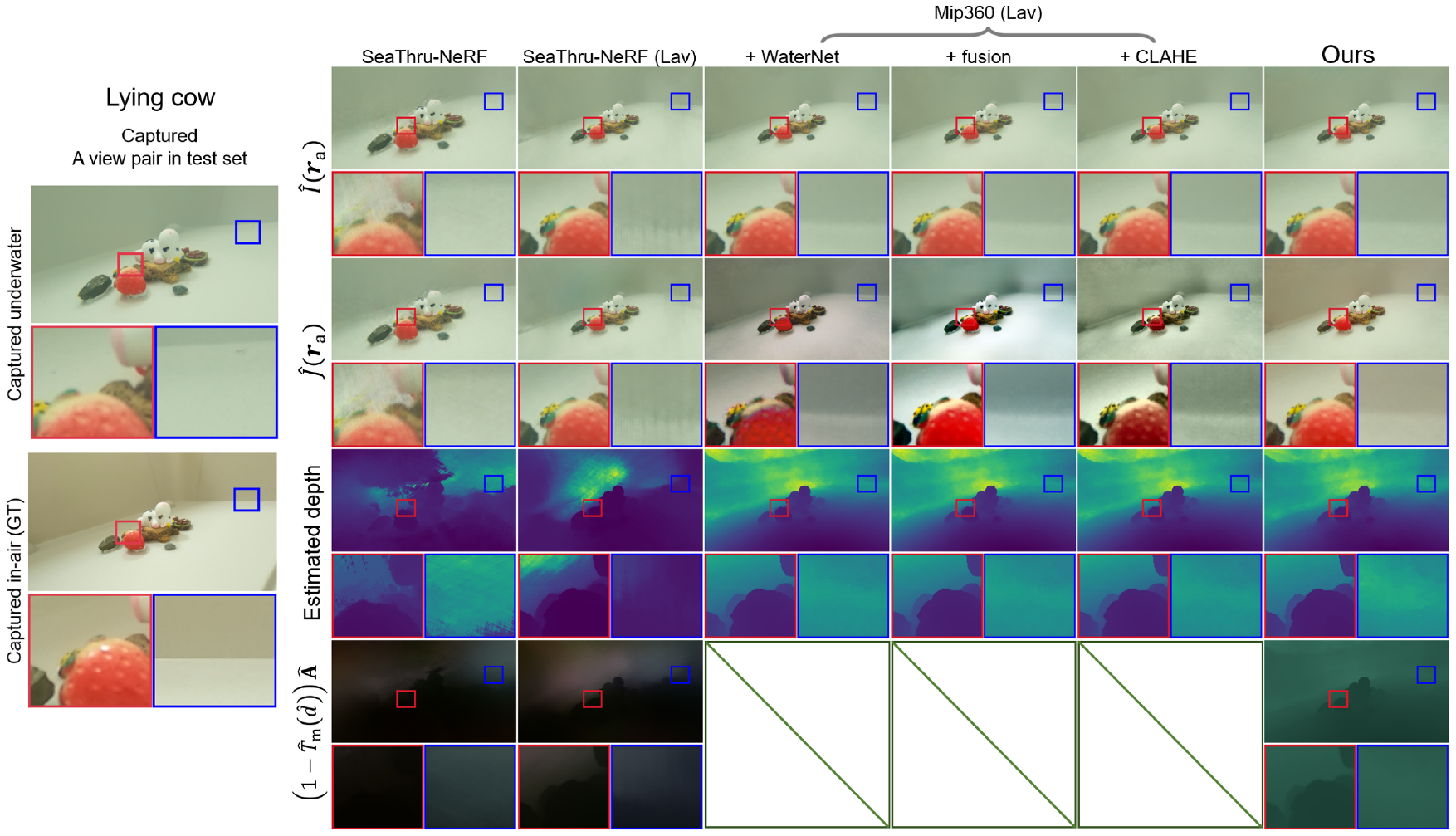}
        \centering{(c) Lying cow}
        \includegraphics[width=1.0\textwidth]{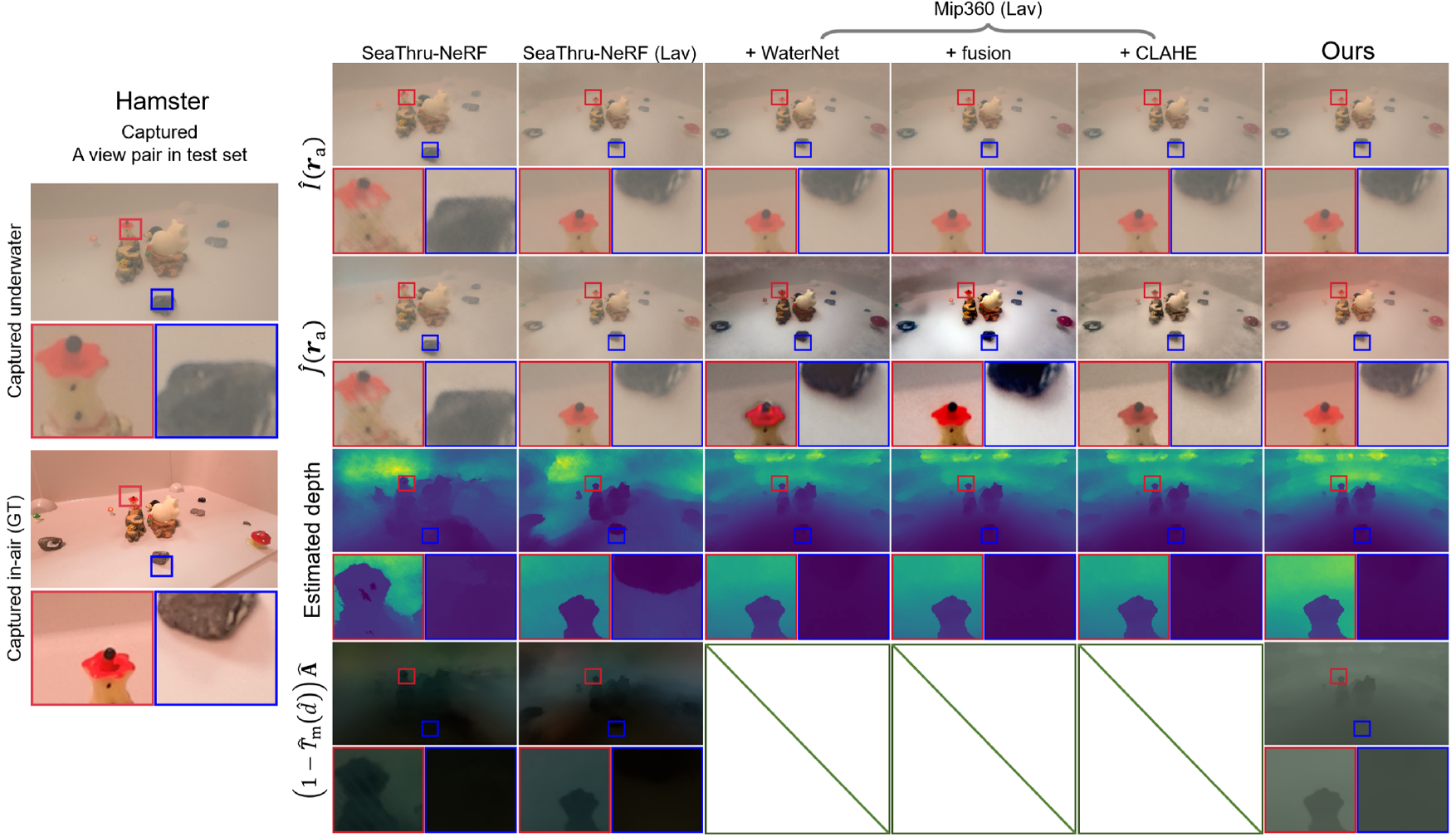}
        \centering{(d) Hamster}
        \vspace{-2mm}
        \caption{Complete qualitative comparison between NeuroPump and other baselines on our benchmark dataset (excluding Totoro, which is already included in the main paper's \autoref{fig:airwater}).
        The 1st row is the underwater image with \textit{geometric rectification only}. Clearly, the original SeaThru-NeRF's result looks like a zoomed-in version of the in-air ground truth, because it cannot rectify refraction. 
        The 2nd row is the underwater image with \textit{both geometric and color rectification (the final target)}. Note that SeaThru-NeRF and SeaThru-NeRF (Lav) cannot accurately rectify underwater color, and the estimated depth (the 3rd row) and back-scatter (the 4th row) are also inferior. Mip360 (Lav)-based methods exhibit differences in brightness and saturation compared to the in-air ground truth. In comparison, our NeuroPump shows clear advantages in all results.}
        \label{fig:suppl_data}
\end{figure*}

    \clearpage 
    
{
    \small
}

\end{document}